\documentclass[10pt, a4paper]{article}
\usepackage{lrec}
\usepackage{multibib}
\newcites{languageresource}{Language Resources}
\usepackage{graphicx}
\usepackage{tabularx}
\usepackage{soul}

\usepackage{flushend}  

\usepackage{epstopdf}
\usepackage[latin1]{inputenc}

\usepackage{amsmath}
\usepackage{hyperref}
\usepackage{xstring}

\title{Open Subtitles Paraphrase Corpus for Six Languages}

\name{Mathias Creutz}

\address{Department of Digital Humanities, Faculty of Arts, University of Helsinki, Finland \\
                  mathias.creutz@helsinki.fi}

\abstract{
This paper accompanies the release of \textit{Opusparcus}, a new paraphrase corpus for six European languages:
German, English, Finnish, French, Russian, and Swedish. The corpus consists of paraphrases, that is, pairs of sentences
in the same language that mean approximately the same thing. The paraphrases are extracted from the OpenSubtitles2016 corpus, which contains
subtitles from movies and TV shows. The informal and colloquial genre that occurs in subtitles makes such data a very
interesting language resource, for instance, from the perspective of computer assisted language learning. For each target language, the Opusparcus data have been 
partitioned into three types of data sets: training, development and test sets. The training sets are large, consisting of millions of sentence pairs, and have 
been compiled automatically, with the help of probabilistic ranking functions. The development and test sets consist of sentence pairs that have been checked manually;
each set contains approximately 1000 sentence pairs that have been verified to be acceptable paraphrases by two annotators.
\\ \newline  }

\begin{document}

\maketitleabstract

\section{Introduction}

This paper introduces the first release of the Opusparcus multilingual corpus of paraphrases \citelanguageresource{opusparcus}. Paraphrases are pairs of phrases in the same language that essentially convey the same meaning, such as \textit{``Have a seat.''} versus \textit{``Sit down.''}. Paraphrase resources have been published earlier, for instance by \newcite{quirk2004emnlp}, \newcite{dolan2004coling}, \newcite{dolan2005iwp}, \newcite{ganitkevitch2013naacl}, \newcite{ganitkevitch2014lrec}, and \newcite{pavlick2015acl}. However, Opusparcus has a few distinctive characteristics.

Firstly, and most importantly, all paraphrases in Opusparcus (\textul{\textbf{Op}}enS\textul{\textbf{u}}btitle\textul{\textbf{sPar}}aphrase\textul{\textbf{C}}orp\textul{\textbf{us}}) consist of movie and TV subtitles extracted from the \textit{OpenSubtitles2016} collection of parallel corpora \cite{lison2016lrec}. Previous paraphrase collections mostly contain fairly formal language in the form of news text and transcripts of parliamentary proceedings. The more \textit{colloquial} language used in subtitles can be a valuable addition, for instance, in computer assisted language learning, to help learners find natural and idiomatic expressions in real-life situations. 

Secondly, in this work the \textit{pivot} language technique introduced by \newcite{bannard2005acl} is applied using multiple pivot languages rather than just one or a few. The technique consists in finding paraphrases in one target language by translating to another, so-called pivot language and then translating back. For example, English \textit{``Have a seat.''} can be translated to French \textit{``Asseyez-vous.''}, which can be translated back to  \textit{``Sit down.''}. Now, a well known fact is that different languages make different distinctions; for instance, the English pronoun \textit{you} corresponds to French \textit{toi} or \textit{vous}, depending on number and degree of politeness. If French paraphrases are extracted using English as a pivot, then the \textit{toi/vous} distinction will typically disappear, such that \textit{``Asseyez-vous.''} and \textit{``Assieds-toi.''} emerge as paraphrases, because they can both be translated as \textit{``Sit down.''}. Whether this is desirable or not depends on the application. However, if multiple pivot languages are used rather than one, more distinctions can be preserved. \newcite{bannard2005acl} use four pivot languages in order to identify English paraphrases. \newcite{denkowski2010wmt} use one, two, or three pivot languages for their five target languages. \newcite{ganitkevitch2014lrec} produce paraphrases for an impressive number of 21 languages, but they limit themselves to using one language, English, as their pivot (in order to be able to use syntactic information, which is available only for English). Opusparcus contains paraphrases in six European languages representing four different language branches: German, English, Swedish (Germanic), French (Romance), Russian (Slavic), and Finnish (Finnic). For each of the six languages, all other five languages are used as pivots. 

Thirdly, simplicity is reflected in several aspects of the work. On one hand, only full sentences, so called sentential paraphrases, are produced, unlike \newcite{ganitkevitch2013naacl}, \newcite{ganitkevitch2014lrec}, and \newcite{pavlick2015acl}, who also extract sub-sentential paraphrases, such as individual word pairs, and include the counts of all such fragments in their reported figures. On the other hand, typically subtitles are fairly short, which makes it easier to evaluate and annotate the paraphrase candidates, unlike the complex sentences in the news data of \newcite{dolan2005iwp}. Furthermore, sub-sentential features or syntactic constraints \cite{callison-burch2008emnlp} are not utilized to assess the likelihood that two sentences are paraphrases. If one favors similar sentence structures, there is a risk to miss some interesting idiomatic variation, such as in \textit{``It's what we do.''} $\leftrightarrow$ \textit{``This is our job.''}. Finally, particular to this work is that paraphrases and scores for ranking paraphrases are \textit{symmetric}. The two phrases are equal, for instance in contrast to the incorporation of fine-grained entailment relations \cite{pavlick2015acl,bowman2015emnlp} and the asymmetric  conditional probabilities used by \newcite{bannard2005acl}. 

The rest of this article is split into two main blocks, followed by some concluding remarks. The data sets and annotation scheme are described in Section~\ref{sec:datasets_annotation} Alternative ranking functions that can be utilized to produce large paraphrase corpora are evaluated in  Section~\ref{sec:ranking}







\begin{table*}[t]
\begin{center}
\begin{tabularx}{\textwidth}{| p{0.11\textwidth} | p{0.4\textwidth} | X |}

      \hline
      Category & Description & Examples \\
      \hline
      Good  \mbox{\textit{``Green''}} & The two sentences can be used in the same situation and essentially ``mean the same thing''. & \mbox{\textit{It was a last minute thing.} $\leftrightarrow$ \textit{This wasn't planned.}} \mbox{\textit{Honey, look.} $\leftrightarrow$ \textit{Um, honey, listen.}} \mbox{\textit{I have goose flesh.}} $\leftrightarrow$ \textit{The hair's standing up on my arms.}\\
      \hline
      \mbox{Mostly good} \mbox{\textit{``Light green''}} & It is acceptable to think that the two sentences refer to the same thing, although one sentence might be more specific than the other one, or there are differences in style, such as polite form versus familiar form. & \mbox{\textit{Hang that up.} $\leftrightarrow$ \textit{Hang up the phone.}} \mbox{\textit{Go to your bedroom.} $\leftrightarrow$ \textit{Just go to sleep.}} \mbox{\textit{Next man, move it.} $\leftrightarrow$ \textit{Next, please.}} \mbox{\textit{Calvin, now what?} $\leftrightarrow$ \textit{What are we doing?}} \mbox{\textit{Good job.} $\leftrightarrow$ \textit{Right, good game, good game.}} \\
      \hline
     \mbox{Mostly bad} \mbox{\textit{``Yellow''}} & There is some connection between the sentences that explains why they occur together, but one would not really consider them to mean the same thing. & \mbox{\textit{Another one?} $\leftrightarrow$ \textit{Partner again?}} \mbox{\textit{Did you ask him?} $\leftrightarrow$ \textit{Have you asked her?}} \mbox{\textit{Hello, operator?} $\leftrightarrow$} \textit{Yes, operator, I'm trying to get to the police.} \\
      \hline
     Bad $\quad$ \mbox{\textit{``Red''}} & There is no obvious connection. The sentences mean different things. & \mbox{\textit{She's over there.} $\leftrightarrow$ \textit{Take me to him.}} \mbox{\textit{All the cons.} $\leftrightarrow$ \textit{Nice and comfy.}} \\
      \hline

\end{tabularx}
\caption{The four annotation categories used, with examples. Each category is also associated with a color, which corresponds to the color of a button in the user interface of the annotation tool.}
\label{tab:annot}
\end{center}
\end{table*}

\section{Data Sets and Annotation Scheme}
\label{sec:datasets_annotation}

OpenSubtitles2016 \cite{lison2016lrec} is a collection of translated movie and TV subtitles from \url{www.opensubtitles.org}. OpenSubtitles2016, which is a subset of the larger OPUS collection\footnote{OPUS (``... the open parallel corpus''): \url{opus.lingfil.uu.se}}, provides a large number of sentence-aligned parallel corpora in 65 languages. When subtitles exist for the same film in multiple languages, then sentence alignments are available for each language pair. For the present work, fifteen such bitexts were used, that is, all language-pair combinations for the six target languages German, English, Finnish, French, Russian and Swedish.

In principle, we work on full sentences only, and thus the terms \textit{sentence} and \textit{phrase} are used fairly interchangeably in this paper. Only one-to-one aligned sentences are used, that is, one sentence in the target language must be aligned with one sentence in the pivot language. There are occasional OCR errors and incorrect sentence segmentation in the data.

For each language, the data have been partitioned into separate training, development, and test sets, based on the release year of the movie; the test sets were extracted from years ending in 4, development sets from years ending in 5, and training sets from the rest. 

Four different categories have been used when annotating sentence pairs. The annotation scheme is illustrated in Table~\ref{tab:annot}. Other annotation schemes exist as well, such as slightly more complex, five-level Likert scales  \cite{callison-burch2008emnlp}.

\subsection{Training Sets}
\label{sec:trainingsets}

The so-called training sets are orders of magnitudes larger than the development and test sets and consist of lists of automatically ranked sentence pairs, where a high rank means a higher probability that the two sentences are paraphrases. The training sets, or subsets of them, are intended to be used freely for any useful purpose.

\begin{figure}[!h]
\begin{center}
\includegraphics[width=0.5\textwidth]{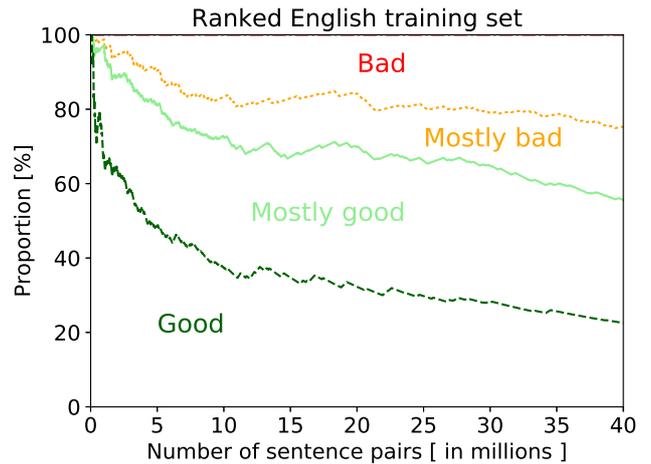} 
\caption{Estimated quality of the English training set. Proportions of the four annotation categories are calculated cumulatively, starting from the most highly ranked sentence pairs (on the left). By picking only highly ranked sentence pairs, one can achieve a high share of ``Good'' and ``Mostly good'' parapahrase candidates. The more data that is included, the more ``Bad'' or ``Mostly bad'' sentence pairs appear in the set.}
\label{fig:engtrain}
\end{center}
\end{figure}

The training sets were produced as follows: First, around 1000 randomly selected sentence pairs were annotated by the author for each of the six languages. Then, an automatic ranking function was applied to all sentence pairs (annotated and unannotated alike), as explained later in Section~\ref{sec:ranking} By extrapolating from the manually annotated data points to the entire set, an estimate of the quality of the training sets can be obtained. The result for English is shown in Figure~\ref{fig:engtrain}.

Another view to the quality of the training sets is provided in Table~\ref{tab:trainingset}, where approximate corpus sizes are given for each of the six languages, at three different accuracy levels.

\begin{table}[!h]
\begin{center}
\begin{tabularx}{\columnwidth}{| X | r | r | r | r | r |}
\hline
Language & 95\% & 90\% & 75\% \\
\hline
German (de) & 590,000 & 1,000,000 & 4,700,000 \\
\hline
English (en) & 1,000,000 & 1,500,000 & 7,000,000 \\
\hline
Finnish (fi) & 480,000 & 640,000 & 3,000,000 \\
\hline
French (fr) & 940,000 & 2,400,000 & 11,000,000 \\
\hline
Russian (ru) & 150,000 & 170,000 & 3,400,000 \\
\hline
Swedish (sv) & 240,000 & 600,000 & 1,400,000 \\
\hline
\end{tabularx}
\caption{Number of phrase pairs in the training sets at three different cut-off points, where 95\%, 90\%, and 75\% of the sentence pairs are estimated to be ``Good'' or ``Mostly good'' paraphrases.}
\label{tab:trainingset}
\end{center}
\end{table}

\subsection{Development and Test Sets}

Whereas the training sets have been produced semi-automatically, the development and test sets consist exclusively of sentence pairs that have been annotated manually. This is to guarantee the high quality of these sets. However, quality comes at the expense of quantity, so the development and test sets are smaller than the training sets. The number of annotations produced for each language are shown i Table~\ref{tab:devandtestsets}. Half of the sentence pairs belong to the development set and the other half to the test set.

\begin{table}[!h]
\begin{center}
\begin{tabularx}{\columnwidth}{| X | r | r |}
\hline
& \multicolumn{1}{c|}{Total number of} & \multicolumn{1}{c|}{``Good'' or ``Mostly} \\ 
Language & \multicolumn{1}{c|}{annotations} & \multicolumn{1}{c|}{good'' paraphrases} \\
\hline
German (de) & 3483 & 2060 \\
\hline
English (en) & 3088 & 1997  \\
\hline
Finnish (fi) & 3703 & 1921 \\
\hline
French (fr) & 3847 & 2004 \\
\hline
Russian (ru) & 4381 & 2088 \\
\hline
Swedish (sv) & 4178 & 1931 \\
\hline
Total & 22,680 & 12,001 \\
\hline
\end{tabularx}
\caption{Total number of manual annotations in the development and test sets combined. Each sentence pair has been annotated independently by two annotators. For each language, approximately 2000 annotated sentence pairs qualify as acceptable paraphrases.}
\label{tab:devandtestsets}
\end{center}
\end{table}

The development sets can be used to refine whatever training algorithms one might want to devise, and the test sets should be used in final evaluations only. The development sets contain only sentence pairs that do not occur in the training sets. The test sets consist of sentence pairs that do not occur in either the training sets or the development sets. 

The sentence pairs to be annotated manually were subject to more rigorous pre-filtering than the sentence pairs in the training sets. In the data, there are many sentences that differ only slightly from each other, such as: \textit{``He is not your friend.''} $\leftrightarrow$ \textit{``He isn't your friend.''}. It would have been a waste of human labor to have such simple and predictable variations annotated manually. Therefore, only pairs of sentences that differ sufficiently from each other are accepted into the development and test sets. The difference is measured using relative edit distance; in general, the edit distance between the two sentences has to be at least 0.4 times the length of the shorter of the sentences (and for very short sentences containing less than 24 characters, the distance threshold is even higher).

Two persons annotated every sentence pair. If the annotators agreed on the category, the annotation was accepted as is. If the annotators disagreed but picked adjacent categories (such as ``Good'' versus ``Mostly good'' or ``Mostly good'' versus ``Mostly bad''), then the annotation was also accepted, but the lower category was assigned (such that ``Mostly good'' and ``Mostly bad'' yields ``Mostly bad''). If there was stronger disagreement between the annotators (such as ``Mostly good'' versus ``Bad''), then the sentence pair was discarded. The annotators were also able to discard a sentence pair, if the language of either sentence was wrong or there were spelling or grammar errors. The number of trashed sentences turned out to be highest for French and Russian: It appears that French orthography is complex and mistakes are fairly common in written text. In the Russian data, some non-Russian Cyrillic as well as Latin characters show up occasionally, apparently because of inaccurate optical character recognition (OCR).

The detailed outcome of the annotation effort is summarized in Table~\ref{tab:devset} for the development sets and Table~\ref{tab:testset} for the test sets.

%

\begin{table*}[t]
\begin{center}
\begin{tabularx}{\textwidth}{| X | r | r | r | r | r | r | r | r | r |}
\hline
& \multicolumn{1}{c|}{Good} & \multicolumn{2}{c|}{Mostly good} & \multicolumn{2}{c|}{Mostly bad} & \multicolumn{2}{c|}{Bad} & \multicolumn{2}{c|}{Discarded} \\
& \multicolumn{1}{c|}{\scriptsize 2 x Green} & \multicolumn{1}{c}{\scriptsize Gr. + Light gr.} & \multicolumn{1}{c|}{\scriptsize 2 x Light green} &  \multicolumn{1}{c}{\scriptsize Light gr. + Yel. } & \multicolumn{1}{c|}{\scriptsize 2 x Yellow} & \multicolumn{1}{c}{\scriptsize Yellow + Red} & \multicolumn{1}{c|}{\scriptsize 2 x Red}  & \multicolumn{1}{c}{\scriptsize Trash} &  \multicolumn{1}{c|}{\scriptsize Disagree} \\
\hline
de & \small 286 \scriptsize (16.4\%) & \small 333 \scriptsize (19.1\%) & \small 394 \scriptsize (22.6\%) & \small 189 \scriptsize (10.9\%) & \small 112 \scriptsize (6.4\%) & \small 100 \scriptsize (5.7\%) & \small 168 \scriptsize (9.6\%) & \small 81 \scriptsize (4.7\%) & \small 79 \scriptsize (4.5\%) \\
\hline
en & \small 409 \scriptsize (26.5\%) & \small 319 \scriptsize (20.7\%) & \small 287 \scriptsize (18.6\%) & \small 105 \scriptsize (6.8\%) & \small 74 \scriptsize (4.8\%) & \small 48 \scriptsize (3.1\%) & \small 213 \scriptsize (13.8\%) & \small 61 \scriptsize (4.0\%) & \small 28 \scriptsize (1.8\%) \\ 
\hline
fi & \small 351 \scriptsize (19.0\%) & \small 268 \scriptsize (14.5\%) & \small 344 \scriptsize (18.6\%) & \small 185 \scriptsize (10.0\%) & \small 135 \scriptsize (7.3\%) & \small 135 \scriptsize (7.3\%) & \small 342 \scriptsize (18.5\%) & \small 36 \scriptsize (1.9\%) & \small 56 \scriptsize (3.0\%) \\
\hline
fr & \small 252 \scriptsize (13.1\%) & \small 337 \scriptsize (17.5\%) & \small 408 \scriptsize (21.2\%) & \small 226 \scriptsize (11.7\%) & \small 207 \scriptsize (10.8\%) & \small 94 \scriptsize (4.9\%) & \small 106 \scriptsize (5.5\%) & \small 229 \scriptsize (11.9\%) & \small 65 \scriptsize (3.4\%) \\
\hline
ru & \small 473 \scriptsize (21.6\%) & \small 337 \scriptsize (15.4\%) & \small 210 \scriptsize (9.6\%) & \small 256 \scriptsize (11.7\%) & \small 155 \scriptsize (7.1\%) & \small 221 \scriptsize (10.1\%) & \small 202 \scriptsize (9.2\%) & \small 185 \scriptsize (8.4\%) & \small 152 \scriptsize (6.9\%) \\
\hline
sv & \small 376 \scriptsize (18.0\%) & \small 303 \scriptsize (14.5\%) & \small 305 \scriptsize (14.6\%) & \small 155 \scriptsize (7.4\%) & \small 86 \scriptsize (4.1\%) & \small 161 \scriptsize (7.7\%) & \small 501 \scriptsize (24.0\%) & \small 105 \scriptsize (5.0\%) & \small 97 \scriptsize (4.6\%) \\
\hline
\end{tabularx}
\caption{Detailed breakdown of the results of the annotation of the development sets. A sentence pair qualifies as a ``good'' paraphrase, when both annotators have chosen the ``good'' category, visualized as a green button in the annotation tool. A sentence pair qualifies as ``mostly good'', when either one annotator has pushed the green button and the other annotator has pushed the light green button or both annotators have chosen the light green button. Similarly, sentence pairs have been categorized as ``mostly bad'' or ``bad'', if both annotators have agreed on the same category or if the annotators ended up pushing adjacent buttons. Sentence pairs were discarded in the following scenarios: The pair was trashed, if at least one of the annotators judged it to contain incorrect spelling or grammar. The sentence pair was also discarded, if the annotators disagreed about the category by more than one step on the four-level scale.}
\label{tab:devset}
\end{center}
\end{table*}

\begin{table*}[t]
\begin{center}
\begin{tabularx}{\textwidth}{| X | r | r | r | r | r | r | r | r | r |}
\hline
& \multicolumn{1}{c|}{Good} & \multicolumn{2}{c|}{Mostly good} & \multicolumn{2}{c|}{Mostly bad} & \multicolumn{2}{c|}{Bad} & \multicolumn{2}{c|}{Discarded} \\
& \multicolumn{1}{c|}{\scriptsize 2 x Green} & \multicolumn{1}{c}{\scriptsize Gr. + Light gr.} & \multicolumn{1}{c|}{\scriptsize 2 x Light green} &  \multicolumn{1}{c}{\scriptsize Light gr. + Yel. } & \multicolumn{1}{c|}{\scriptsize 2 x Yellow} & \multicolumn{1}{c}{\scriptsize Yellow + Red} & \multicolumn{1}{c|}{\scriptsize 2 x Red}  & \multicolumn{1}{c}{\scriptsize Trash} &  \multicolumn{1}{c|}{\scriptsize Disagree} \\
\hline
de & \small 303 \scriptsize (17.4\%) & \small 333 \scriptsize (19.1\%) & \small 411 \scriptsize (23.6\%) & \small 177 \scriptsize (10.2\%) & \small 116 \scriptsize (6.7\%) & \small 85 \scriptsize (4.9\%) & \small 161 \scriptsize (9.2\%) & \small 77 \scriptsize (4.4\%) & \small 78 \scriptsize (4.5\%) \\
\hline
en & \small 450 \scriptsize (29.1\%) & \small 273 \scriptsize (17.7\%) & \small 259 \scriptsize (16.8\%) & \small 97 \scriptsize (6.3\%) & \small 56 \scriptsize (3.6\%) &  \small 64 \scriptsize (4.1\%) & \small 246 \scriptsize (15.9\%) & \small 59 \scriptsize (3.8\%) & \small 40 \scriptsize (2.6\%) \\ 
\hline
fi & \small 376 \scriptsize (20.3\%) & \small 244 \scriptsize (13.2\%) & \small 338 \scriptsize (18.3\%) & \small 179 \scriptsize (9.7\%) & \small 138 \scriptsize (7.5\%) & \small 121 \scriptsize (6.5\%) & \small 353 \scriptsize (19.1\%) & \small 60 \scriptsize (3.2\%) & \small 42 \scriptsize (2.3\%) \\
\hline
fr & \small 261 \scriptsize (13.6\%) & \small 337 \scriptsize (17.5\%) & \small 409 \scriptsize (21.3\%) & \small 206 \scriptsize (10.7\%) & \small 204 \scriptsize (10.6\%) & \small 124 \scriptsize (6.4\%) & \small 133 \scriptsize (6.9\%) & \small \,\,184 \scriptsize (9.6\%) & \small 65 \scriptsize (3.4\%) \\
\hline
ru & \small 462 \scriptsize (21.1\%) & \small 351 \scriptsize (16.0\%) & \small 255 \scriptsize (11.6\%) & \small 223 \scriptsize (10.2\%) & \small 151 \scriptsize (6.9\%) & \small 225 \scriptsize (10.3\%) & \small 188 \scriptsize (8.6\%) & \small 189 \scriptsize (8.6\%) & \small 146 \scriptsize (6.7\%) \\
\hline
sv & \small 379 \scriptsize (18.1\%) & \small 312 \scriptsize (14.9\%) & \small 256 \scriptsize (12.3\%) & \small 173 \scriptsize (8.3\%) & \small 107 \scriptsize (5.1\%) & \small 147 \scriptsize (7.0\%) & \small 527 \scriptsize (25.2\%) & \small 120 \scriptsize (5.7\%) & \small 68 \scriptsize (3.3\%) \\
\hline
\end{tabularx}
\caption{Detailed breakdown of the results of the annotation of the test sets. Exactly the same procedure was applied as for the development sets. Annotators were unaware of which set a particular sentence pair belonged two; in fact, most annotators were unaware of the existence of separate development and test sets.}
\label{tab:testset}
\end{center}
\end{table*}

\section{Automatic Ranking of Paraphrase Candidates}
\label{sec:ranking}

For the data sets that are intended to be used as training sets, a number of ranking schemes have been tested in order to identify paraphrases. Five of the ranking schemes are presented below, followed by a description how these approaches were evaluated.
In the examples, English is used as our target language, and we are looking for English paraphrases. In the actual experiments, English was just one of the languages, and the same procedure was carried out for German, Finnish, French, Russian, and Swedish, as well.

\subsection{Conditional Probability}

\newcite{bannard2005acl} propose a conditional paraphrase probability $P(e_2 | e_1)$ as the probability that the English phrase $e_1$ is translated to a foreign phrase $f_i$, which in turn is translated back into another English phrase $e_2$. Since there are typically multiple possible foreign translations, we need to marginalize over the different possible $f_i$:
\begin{equation}
\label{eq:condprob}
P(e_2 | e_1) = \sum_i P(e_2 | f_i) P(f_i | e_1)
\end{equation}

This ranking formula tends to assign high ranks to phrase pairs, where $e_1$ is more specific than $e_2$. For instance, consider the case, where $e_1$ is \textit{``I was taken from my family when I was a boy.''} and $e_2$ is \textit{``I was taken from my family.''}. In the English-French parallel corpus, both English phrases have been aligned with the French phrase $f_1$: \textit{``On m'a enlevé à ma famille.''}. However, $e_1$ occurs aligned against $f_1$ only once, whereas $e_2$ 21 times. Thus, $P(f_1 | e_1)$ is high~(=1), because $e_1$ is always translated as $f_1$. Also, $P(e_2 | f_1)$ is high~(= 21/22), because $f_1$ is almost always translated as $e_2$.

This tendency produces numerous errors, when there are occasional misaligned phrases in the corpus, such as in: \textit{``We're staying in the army.''} $\rightarrow$ \textit{``Aah.''}, where \textit{``We're staying in the army.''} has been aligned against French \textit{``Aah.''} once, which in turn has been aligned with English \textit{``Aah.''} 1401 times.

\subsection{Joint Probability}

Instead of a conditional probability, which is asymmetric, one can use the corresponding joint probability, which includes a prior probability, and is symmetric. Thus, the probability of $e_1$ being a paraphrase of $e_2$ is the same as the probability of $e_2$ being a paraphrase of $e_1$:
\begin{equation}
\label{eq:jointprob}
P(e_1, e_2) = P(e_2 | e_1) P(e_1) = P(e_1 | e_2) P(e_2)
\end{equation}
$P(e_2 | e_1)$ and $P(e_1 | e_2)$ are calculated as in Equation~\eqref{eq:condprob}, and $P(e_1)$ and $P(e_2)$ are the (prior) probabilities of the phrases, which are simply estimated as relative frequencies over all sentences in the corpus.

Now, at the top of the ranking, we find pairs consisting of frequently used phrases: \textit{``Yes.''} $\leftrightarrow$ \textit{``Yeah.''}, \textit{``Of course.''} $\leftrightarrow$ \textit{``Sure.''}, \textit{``Hello.''} $\leftrightarrow$ \textit{``Good morning.''}, \textit{``Are you okay?'}' $\leftrightarrow$  \textit{``Are you all right?''}.

A few spurious phrase pairs also score high, where it appears that two frequent phrases might have found a common translation mostly by chance, by the fact that they occur frequently in general: \textit{``You 're welcome.''} $\leftrightarrow$ \textit{``Sure.''}, \textit{``Yeah.''} $\leftrightarrow$ \textit{``I am.''}, \textit{``Hi.''} $\leftrightarrow$ \textit{``Goodbye.''}, \textit{``I do.''} $\leftrightarrow$ \textit{``I know.''}

\subsection{Pointwise Mutual Information}

Pointwise Mutual Information (PMI) divides the joint probability by the probability that the two phrases $e_1$and $e_2$ occur independently. Thus, PMI penalizes phrase pairs that co-occur mostly by chance, by the fact that they occur frequently in general:
 \begin{equation}
\label{eq:pmi}
\begin{split}
\text{pmi}(e_1; e_2) & = \log \frac{P(e_1, e_2)}{P(e_1)P(e_2)} \\ 
& = \log \frac{P(e_2 | e_1)}{P(e_2)} = \log \frac{P(e_1 | e_2)}{P(e_1)}
\end{split}
\end{equation}

This scoring favors phrase pairs $e_1$ and $e_2$ that have a limited set of translations $f_i$, such that $e_1$ and $e_2$ are not aligned with phrases other than $f_i$, and $f_i$ are not aligned with other phrases than  $e_1$ and $e_2$. For instance, the phrases \textit{``You sound a little homesick.''} $\leftrightarrow$ \textit{``Do you miss being home?''} have a common French translation \textit{``Vous avez le mal du pays ?''}, which occurs twice in the corpus, aligned once against each of the two English phrases.

However, similarly to the conditional probability in Equation~\eqref{eq:condprob}, PMI is sensitive to misaligned, infrequent sentences. The phrase pair  \textit{``Lost the phone now.''} $\leftrightarrow$ \textit{``I'm from the agency.''} scores high because \textit{``Lost the phone now.''} has been misaligned against French \textit{``Je viens de l'agence.''}, which occurs only twice.
 
\subsection{Joint Probability and PMI Combined} 

Our experiments show that rather than using joint probability~\eqref{eq:jointprob} or PMI~\eqref{eq:pmi} in isolation, we obtain a better ranking by multiplying the two together:  
\begin{equation}
\label{eq:jointprobtimespmi}
P(e_1, e_2) \cdot \text{pmi}(e_1; e_2) = P(e_1, e_2) \log \frac{P(e_1, e_2)}{P(e_1)P(e_2)}
\end{equation} 

This leverages the strengths and alleviates the short-comings of the two approaches.

\subsection{Multiple Multilingual Parallel Corpora}

The four formulae presented above, \eqref{eq:condprob}, \eqref{eq:jointprob}, \eqref{eq:pmi}, and \eqref{eq:jointprobtimespmi}, easily generalize beyond bilingual parallel corpora. One can simply concatenate all parallel corpora, such that English is kept on one side and the other (pivot) languages on the other side. All frequencies and probabilities are then calculated over the merged bitext as a whole.

Interestingly, another approach can produce better results. PMI in Equation~\eqref{eq:pmi} may rank rare phrase pairs very high, and in case of misalignments, these pairs are unreliable. However, if a rare phrase pair ranks high in multiple bitexts, then this seems to signal much higher confidence. In order not to lose the information that a phrase pair emerges in multiple different corpora, rather than merging the parallel corpora into one, we can keep them separate. We then compute PMI scores separately for each bitext (English-German, English-Finnish, English-French, English-Russian, and English-Swedish). To obtain a combined score, we compute the sum of the PMI values obtained from each different corpus (English vs. pivot language $\mathcal{L}_i$): 
\begin{equation}
\label{eq:pmisum}
\sum_i \text{pmi}(e_1; e_2 | \mathcal{L}_i) = \sum_i \log \frac{P(e_1, e_2 | \mathcal{L}_i)}{P(e_1 | \mathcal{L}_i)P(e_2 | \mathcal{L}_i)}
\end{equation}
The probabilities are calculated exactly as previously. The notation merely highlights the fact that every value is conditioned on alignments between English and a specific pivot language $\mathcal{L}_i$.

Since the number of languages is constant, the sum in~\eqref{eq:pmisum} can also be interpreted as the average PMI across pivot languages.

\subsection{Evaluation of Ranking Schemes}

A symmetric score is desired in this work, and therefore the conditional probability in Equation~\eqref{eq:condprob} cannot be used as such. However, one could obtain a symmetric score by combining $P(e_2 | e_1)$ and $P(e_1 | e_2)$ in some way, such as taking the minimum, maximum or average value. In practice, this would make this score behave fairly similarly as the more elegantly formulated PMI in Equation~\eqref{eq:pmi}, so the conditional probability scheme was not investigated further.

This leaves us with the four remaining schemes in Equations~\eqref{eq:jointprob}, \eqref{eq:pmi}, \eqref{eq:jointprobtimespmi}, and \eqref{eq:pmisum}. They were compared with the help of a set of phrase pairs that were drawn randomly from the training set and annotated manually, as described in Section~\ref{sec:trainingsets} The training set is then reordered using the ranking scheme to be tested. Depending on the ranking scheme, the manually annotated phrase pairs will appear at different ranks in the full, ordered collection. An ideal ranking scheme will place the phrase pairs that are true paraphrases at the head of the ordering and the phrase pairs that are not paraphrases at the tail.

The results were then plotted as in Figure~\ref{fig:engtrain} and compared visually. Across the six languages, the results were consistent: the best performing rankings were PMI summed over multiple corpora~\eqref{eq:pmisum} followed by joint probability mutliplied by PMI~\eqref{eq:jointprobtimespmi}. The types of phrase pairs that rank high are different in both cases: the former favors less frequent, more specific phrase pairs, such as \textit{``It was a difficult and long delivery.''} $\leftrightarrow$ \textit{``The delivery was difficult and long.''}, whereas the latter favors frequent, less informative phrase pairs, such as: \textit{``Excuse me.''} $\leftrightarrow$ \textit{``I'm sorry.''}. PMI summed over multiple corpora, in Equation~\eqref{eq:pmisum}, was judged to be the best ranking function. The final training sets were produced using this particular ranking.


\section{Conclusion}


Paraphrase extraction from movie subtitle data has been described in this paper. Six languages were included in this initial phase, but there is 
no principal reason why not more of the 65 languages in the OpenSubtitles2016 collection could be exploited. As there is considerable manual annotation effort involved, crowdsourcing could be considered; see, for instance, \newcite{tschirsich2013lawacl}.

Another improvement could be to reduce the number of OCR errors that still occur in the data.

%
%
%
%
%

\section*{Acknowledgments}

The following people have participated in the annotation effort: Thomas de Bluts, Aleksandr Semenov, Eetu Sjöblom, Mikko Aulamo, Olivia Engström, Janine Siewert, Carola Carpentier, Svante Creutz, Yves Scherrer, Anders Ahlbäck, Sami Itkonen, Riikka Raatikainen, Kaisla Kajava, Tiina Koho, Oksana Lehtonen, Sharid Loáiciga Sánchez, Tatiana Batanina, and the author himself. The annotation tool was implemented by Mikko Aulamo. The author is grateful to Mikko and the annotators for their very valuable contributions.

The author would also like to thank professor Jörg Tiedemann together with the anonymous reviewers for their valuable comments and suggestions as well as the Academy of Finland for the financial support of the annotations through Project 314062 in the ICT 2023 call on Computation, Machine Learning and Artificial Intelligence.

Furthermore, warm thanks go to Mietta Lennes and the Language Bank of Finland (Kielipankki) for publishing and hosting Opusparcus as part of their corpus collections.

\section{Bibliographical References}
\label{main:ref}

\bibliographystyle{lrec}
\bibliography{creutz2018lrec}

\section{Language Resource References}
\label{lr:ref}
\bibliographystylelanguageresource{lrec}
\bibliographylanguageresource{creutz2018lrec}

\end{document}